\pdfoutput=1
\documentclass[11pt]{article}

\usepackage{booktabs}
\usepackage{mdframed}

\usepackage[final]{acl}

\usepackage{times}
\usepackage{latexsym}

\usepackage[T1]{fontenc}

\usepackage[utf8]{inputenc}

\usepackage{csquotes}

\usepackage{microtype}

\usepackage{inconsolata}

\usepackage{graphicx}

\usepackage{amsmath,soul}
\usepackage{soulpos}
\usepackage[dvipsnames]{xcolor}
\usepackage{mdframed}

\usepackage[absolute,showboxes,overlay]{textpos}

\title{AIxcellent Vibes at GermEval 2025 Shared Task on Candy Speech Detection: Improving Model Performance by Span-Level Training}

\author{
 \textbf{Christian Rene Thelen\textsuperscript{1,3}},
 \textbf{Patrick Gustav Blaneck\textsuperscript{1,4}},
 \textbf{Tobias Bornheim\textsuperscript{5}},
 \\
 \textbf{Niklas Grieger\textsuperscript{1,2,6}},
 \textbf{Stephan Bialonski\textsuperscript{1,2}}
\\
\\
 \textsuperscript{1}Department of Medical Engineering and Technomathematics\\ \textsuperscript{2}Institute for Data-Driven Technologies\\ FH Aachen University of Applied Sciences, Jülich, Germany\\
 \\
 \textsuperscript{3}Academic and Research Department Engineering Hydrology\\ \textsuperscript{4}IT Center\\ RWTH Aachen University, Aachen, Germany\\
 \\
 \textsuperscript{5}Department for Data Science and AI\\ ORDIX AG, Paderborn, Germany\\
 \\
 \textsuperscript{6}Department of Information and Computing Sciences\\ Utrecht University, Utrecht, The Netherlands
\\
 \small{
   \textbf{Correspondence:} \href{mailto:bialonski@fh-aachen.de}{bialonski@fh-aachen.de}
 }
}

\date{}

\begin{document}
\maketitle
\begin{abstract}
    Positive, supportive online communication in social media (\enquote{candy speech}) has the potential to foster civility, yet automated detection of such language remains underexplored, limiting systematic analysis of its impact.
    We investigate how candy speech can be reliably detected in a 46k-comment German YouTube corpus by monolingual and multilingual language models, including GBERT, Qwen3 Embedding, and XLM-RoBERTa.
    We find that a multilingual XLM-RoBERTa-Large model trained to detect candy speech at the span level outperforms other approaches, ranking first in both binary (positive F1: 0.8906) and categorized span-based detection (strict F1: 0.6307) subtasks at the \emph{GermEval 2025 Shared Task on Candy Speech Detection}.
    We speculate that span-based training, multilingual capabilities, and emoji-aware tokenizers improved detection performance.
    Our results demonstrate the effectiveness of multilingual models in identifying positive, supportive language.
\end{abstract}

\begin{textblock*}{20cm}[0.5,0](10.5cm,27.8cm)
  \centering
  \small
  This work was peer-reviewed and published in the Proceedings of the 21st Conference on Natural Language Processing (KONVENS 2025): Workshops, available at \url{https://aclanthology.org/2025.konvens-2.33/}. 
  Please cite as: Thelen CR, Blaneck PG, Bornheim T, Grieger N, and Bialonski S. 2025. AIxcellent Vibes at GermEval 2025 Shared Task on Candy Speech Detection: Improving Model Performance by Span-Level Training. In Proceedings of the 21st Conference on Natural Language Processing (KONVENS 2025): Workshops, pages 398–403, Hannover, Germany.
\end{textblock*}

\section{Introduction}
\label{sec:introduction}

User comments on social media can significantly shape the perceptions of discourse participants and alter the trajectory of debates \cite{Kubin2024}.
While research has been dedicated to detecting negative user contributions, such as hate speech \cite{Albladi2025}, corresponding mitigation strategies like counterspeech \cite{Bonaldi2024}, positive and empowering expressions, known as candy speech (German: \enquote{Flausch}; lit. \enquote{fluff} \cite{Clausen2025}) or hope speech \cite{Chakravarthi2022b} remain underexplored.
To facilitate systematic analysis of candy speech, automated annotation tools are desirable that allow the detection of such speech events in large datasets at a level of granularity that would be impractical for human annotators to achieve at scale.

Detection methods for candy speech can build upon established natural language processing (NLP) techniques, ranging from traditional rule-based systems to contemporary data-driven approaches based on transformer models pretrained on large text corpora~\cite{Tucudean2024}.
Encoder-only transformer architectures, such as BERT \cite{Devlin2018} or XLM-RoBERTa \cite{Conneau2020}, transform text tokens into embedding vectors and have demonstrated strong performance across various NLP classification and regression tasks.
Specifically, monolingual variants like the German BERT model GBERT \cite{Chan2020} have achieved competitive results on German-language NLP tasks \cite{Bornheim2021,Blaneck2022}.
Decoder-only transformers, including GPT-4o \cite{GPT4-2023}, DeepSeek-V3 \cite{DeepSeek2024}, and Qwen3 \cite{Yang2025}, traditionally used for text generation, have also shown promise in annotation tasks through zero-shot, few-shot, and fine-tuning approaches \cite{Bornheim2024}.
Recent advancements, such as GPT-based embedding models like Qwen3 Embedding, further enhance these capabilities by generating embeddings conditioned on specific task-related instructions \cite{Zhang2025}.

In this contribution, we investigate the performance of monolingual (GBERT) and multilingual (Qwen3 Embedding, XLM-RoBERTa) language models in detecting candy speech within a novel dataset of German YouTube comments \cite{Clausen2025}.
Our experiments demonstrate that XLM-RoBERTa-Large, when trained to detect spans of various candy speech types using the BIO tagging scheme \cite{Ramshaw1995}, outperforms all other approaches, including binary classifiers designed solely to distinguish candy from non-candy speech.
We share our implementation online.\footnote{\url{https://github.com/dslaborg/germeval2025}}

\section{Data and Tasks}
\label{sec:data_task}

The \emph{GermEval 2025 Shared Task on Candy Speech Detection} consisted of two subtasks.
For the first subtask, participants developed a binary classification model to predict whether a YouTube comment contained candy speech.
The second subtask required developing a model that could identify candy speech spans of different types (i.e., \emph{positive feedback}, \emph{compliment}, \emph{affection declaration}, \emph{encouragement}, \emph{gratitude}, \emph{agreement}, \emph{ambiguous}, \emph{implicit}, \emph{group membership}, \emph{sympathy}).

\begin{figure}
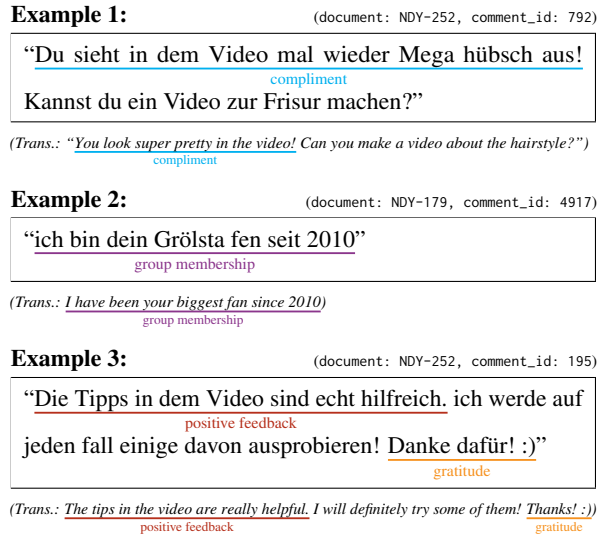


    \ulposdef{\ulspanaux}{%
        \normalfont$\underset{\textcolor{\spancolor}{\text{\spanlabel}}}{\textcolor{\spancolor}{\rule[-.7ex]{\ulwidth}{.7pt}}}$%
    }

    \newcommand{\ulspan}[2]{%
        \def\saveulspan{#1}%
        \def\spanlabel{%
            \ifcase\numexpr#1 \relax
                positive feedback%
            \or compliment%
            \or affection declaration%
            \or encouragement%
            \or gratitude%
            \or agreement%
            \or ambiguous%
            \or implicit%
            \or group membership%
            \or sympathy%
            \fi}%
        \def\spancolor{%
            \ifcase\numexpr#1 \relax
                BrickRed%
            \or ProcessBlue%
            \or Peach%
            \or CarnationPink%
            \or BurntOrange%
            \or Magenta%
            \or LimeGreen%
            \or Turquoise%
            \or Fuchsia%
            \or BlueViolet%
            \fi}%
        \ulspanaux{#2}%
    }
    \small
    \textbf{Example 1:}%
    \hfill%
    {\tiny(\texttt{%
            document: NDY-252, comment\_id: 792%
        })}

    \vspace{-0.75em}

    \begin{mdframed}[
            leftmargin=0pt,
            rightmargin=0pt,
            innertopmargin=0.5em,
            innerbottommargin=0.5em,
            innerrightmargin=0.5em,
            innerleftmargin=0.5em,
        ]
        \enquote{%
            \ulspan{1}{Du sieht in dem Video mal wieder Mega hübsch aus!} Kannst du ein Video zur Frisur machen?%
        }
    \end{mdframed}

    \vspace{-0.5em}

    {\tiny{\emph{(Trans.: %
                \enquote{%
                    \ulspan{1}{You look super pretty in the video!} Can you make a video about the hairstyle?%
                })}}}

    \vspace{0.75em}

    \textbf{Example 2:}%
    \hfill%
    {\tiny(\texttt{%
            document: NDY-179, comment\_id: 4917%
        })}

    \vspace{-0.75em}

    \begin{mdframed}[
            leftmargin=0pt,
            rightmargin=0pt,
            innertopmargin=0.5em,
            innerbottommargin=0.5em,
            innerrightmargin=0.5em,
            innerleftmargin=0.5em,
        ]
        \enquote{%
            \ulspan{8}{ich bin dein Grölsta fen seit 2010}%
        }

        \vspace{-0.25em}
    \end{mdframed}

    \vspace{-0.5em}

    {\tiny{\emph{(Trans.: %
                \ulspan{8}{I have been your biggest fan since 2010}%
                )}}}

    \vspace{0.75em}

    \textbf{Example 3:}%
    \hfill%
    {\tiny(\texttt{%
            document: NDY-252, comment\_id: 195%
        })}

    \vspace{-0.75em}

    \begin{mdframed}[
            leftmargin=0pt,
            rightmargin=0pt,
            innertopmargin=0.5em,
            innerbottommargin=0.5em,
            innerrightmargin=0.5em,
            innerleftmargin=0.5em,
        ]
        \enquote{%
            \ulspan{0}{Die Tipps in dem Video sind echt hilfreich.} ich werde auf jeden fall einige davon ausprobieren! \ulspan{4}{Danke dafür! :)}%
        }

        \vspace{-0.25em}
    \end{mdframed}

    \vspace{-0.5em}

    {\tiny{\emph{(Trans.: %
                \ulspan{0}{The tips in the video are really helpful.} I will definitely try some of them! \ulspan{4}{Thanks! :)}%
                )}}}

    \caption{
        Examples of span annotations in the training data.
        Each comment is identified by a \texttt{document} and a \texttt{comment\_id}.
        The spans are color-coded and labeled with the corresponding candy speech type.
    }
    \label{fig:span_example}
\end{figure}

The provided dataset was split into a public training set and a blind test set used to evaluate the final Shared Task submissions.
The training set contained 37,057 annotated comments from the comment threads of eleven YouTube videos.
In contrast, the test set contained 9,229 comments from the comment threads of five YouTube videos not included in the training set.
Each comment included a \texttt{document} referencing its source video, a \texttt{comment\_id} that was incremented for each comment, and the comment's text.
The training set also included annotations on whether a comment was candy speech, as well as the spans of different candy speech types as triplets for each span (start character index, end character index, type).
See \autoref{fig:span_example} for examples of span annotations in the training data.
Although many comments included multiple spans (an average of 1.47 per comment), there were only 301 instances of overlapping spans, accounting for roughly 1.9~\% of all spans.

\section{Methods}
\label{sec:methods}

\subsection{Preprocessing and Data Split}
\label{ssec:preprocessing_data_split}

We removed 3,829 comments from the training data as duplicates with identical text and labels.
Furthermore, we identified 99 comments with identical text but differing labels, which we chose to retain.

We used two different data splits for model exploration and final training, respectively.
During model exploration, we split the provided training dataset into five folds for cross-validation, which were stratified by the binary candy speech label or first span type for Subtask 1 and Subtask 2, respectively.
For our submissions to the competition, we recombined the cross-validation folds and retrained the final models on the entire training dataset.
In both model exploration and final training, if not otherwise specified, we reserved 10~\% of the training data for early stopping.

\subsection{Models}
\label{ssec:models}

We based our detection systems on three different open-source pretrained language models (LMs), each extended with task-specific classifiers to address the two Subtasks.

Qwen3~Embedding \cite{Zhang2025} is a family of text embedding models derived from the Qwen3 foundation models \cite{Yang2025}.
This model family contains models of various sizes (0.6B, 4B, 8B) and achieves state-of-the-art performance on tasks like multilingual retrieval.
Their multilingual capabilities and ability to handle a broad range of input tokens (e.g., emojis) make the models well suited for processing internet-style text corpora, such as the dataset used in the GermEval 2025 Shared Task.

GBERT \cite{Chan2020} is a German-language transformer model developed by deepset, based on the BERT architecture \cite{Devlin2018}, and pretrained on a large corpus of German text.
The model is optimized for a variety of downstream NLP tasks in German, including classification, named entity recognition, and semantic similarity.
Its monolingual training makes GBERT particularly effective for domain-specific or informal German texts and has been a popular architecture in previous GermEval Shared Tasks~\cite{Bornheim2021,Blaneck2022,Geiss2024}.

XLM-RoBERTa \cite{Conneau2020} is a multilingual transformer model developed by Facebook AI, built upon the RoBERTa architecture and trained on over 2~TB of filtered CommonCrawl data across 100 languages.
It achieves strong cross-lingual performance, often on par with, and in some cases even surpassing, monolingual models on language-specific tasks.

For Subtask 1, a binary classification task, we explored two approaches: (i) using Qwen3-Embedding-8B\footnote{\url{https://huggingface.co/Qwen/Qwen3-Embedding-8B}} as a feature extractor in combination with a support vector machine (SVM) classifier, and (ii) extending GBERT-Large\footnote{\url{https://huggingface.co/deepset/gbert-large}} with a multi-layer perceptron (MLP) classifier.
In the latter approach, the MLP was applied to the last hidden state of the CLS token and consisted of a linear layer with hidden size 1024, followed by a Tanh activation function, dropout with a rate of 0.1, and a final linear layer with two output units and a Softmax activation function for classification.

We addressed the span detection task in Subtask 2 with a token-wise BIO tagging approach~\cite{Ramshaw1995}.
Since the dataset contained spans for ten different candy speech types, we defined separate B (beginning) and I (inside) labels for each type, resulting in a total of 21 labels (including the O (outside) label for non-candy speech).
In our approach, we appended a linear classification layer with 21 output units to both GBERT-Large and XLM-RoBERTa-Large\footnote{\url{https://huggingface.co/FacebookAI/xlm-roberta-large}}.
The linear layer was applied to the last hidden state of each token to predict whether the token was part of a span of a specific candy speech type.

\begin{table*}
    \centering
    \small
    \begin{tabular}{lcc}
        \toprule
        & \textbf{Subtask 1}     & \textbf{Subtask 2} \\
        \textbf{Approach}                                & \textbf{Positive F1}   & \textbf{Strict F1} \\
        \midrule
        \emph{Fine-tuning LMs for Spans}                 &                        &                    \\
        \quad \emph{Basic Postprocessing}                &                        &                    \\
        \qquad GBERT-Large                               & 0.903 (0.004)          & 0.731              \\
        \qquad XLM-RoBERTa-Large$^{*}$                   & \textbf{0.913} (0.002) & \textbf{0.747}     \\
        \quad \emph{Extended Postprocessing}             &                        &                    \\
        \qquad GBERT-Large$^{*}$                         & \---                   & 0.739              \\
        \qquad XLM-RoBERTa-Large                         & \---                   & 0.742              \\
        \midrule
        \emph{Training SVM for Binary Classification}    &                        &                    \\
        \quad Qwen3-Embedding-8B$^{*}$                   & 0.901 (0.006)          & \---               \\
        \emph{Fine-tuning LMs for Binary Classification} &                        &                    \\
        \quad GBERT-Large                                & 0.887 (0.004)          & \---               \\
        \bottomrule
    \end{tabular}
    \caption{
        Validation scores obtained for different modeling approaches.
        For Subtask 1, scores are reported as average (standard deviation) over the cross-validation folds.
        For Subtask 2, scores were obtained on the early stopping set of the final retraining before submission.
        Approaches marked with $^{*}$ denote those submitted to the Shared Task.
    }
    \label{tab:results_val}
\end{table*}

\subsection{Training}
\label{ssec:training}

In our first approach to Subtask 1, we extracted embeddings from a frozen Qwen3-Embedding-8B model and used them as input features for an SVM with a radial basis function (RBF) kernel.
The SVM hyperparameters were selected using the cross-validation setup described in Section \ref{ssec:preprocessing_data_split} with a grid search over $C \in [1, 100]$ and $\gamma \in [10^{-8}, 6 \cdot 10^{-4}]$.
Calculating the F1 score of the positive candy speech class (\emph{positive F1}), we found the optimal hyperparameters of the SVM to be $C = 5$ and $\gamma = 2 \cdot 10^{-4}$.
Since the SVM does not require early stopping, we trained the model on the entire training dataset without the need for a separate early stopping set.

Our second approach to Subtask 1 involved fine-tuning a GBERT-Large model together with an MLP classifier.
Different from our first approach, we oversampled the candy speech class in the training data up to a ratio of 1:1 with the non-candy speech class by sampling with replacement from the candy speech class.
Using the resampled data, we trained the model for five epochs to minimize the cross-entropy loss with the AdamW optimizer, a batch size of 32, a weight decay of 0.01, and gradient clipping to an L2 norm of 1.0.
The learning rate increased from 0 to $5 \cdot 10^{-5}$ during a warmup phase spanning the first 30~\% of the total gradient steps, after which it declined linearly back to 0.
We evaluated the model on the early stopping set using the positive F1-score every 44 gradient steps, and stopped training if the score did not improve for 64 consecutive evaluations.

In Subtask 2, we fine-tuned both GBERT-Large and XLM-RoBERTa-Large models, each extended with a linear layer as classification head.
The training setup was similar to the one used for our second approach in Subtask 1, except for the peak learning rate of $2 \cdot 10^{-5}$, the training duration of 20 epochs with evaluations every 40 gradient steps, an early stopping patience of 87 evaluations, and the length of the warmup phase of 200 and 500 gradient steps for GBERT and XLM-RoBERTa, respectively.
To evaluate model performance, we calculated the strict match F1-score (\emph{strict F1}), which counts two spans as matching if they have the same start and end position and the same label.

\subsection{Postprocessing}
\label{ssec:postprocessing}

For Subtask~2, we defined two sets of postprocessing steps to parse predicted BIO-tokens into the required span format.
In the first, \emph{basic} set of postprocessing steps, we parsed all valid spans consisting of a B-token and optional following I-tokens of the same candy speech type into a span start, span end, and span label.
Spans that began with an I-token instead of a B-token were discarded.
Additionally, I-tokens whose types did not match the preceding B- or I-token were also discarded.

The second, \emph{extended} set of postprocessing steps included the basic set and ensured that spans did not begin or end within a word (e.g., before a subword token like \enquote{\#\#en}).
In such cases, we included the subword token in the span and did not initiate a new span, even if it was marked as a B-token.

\section{Results}
\label{sec:results}

\begin{table*}
    \centering
    \small
    \begin{tabular}{lcc}
        \toprule
        & \textbf{Subtask 1}   & \textbf{Subtask 2} \\
        \textbf{Approach}                             & \textbf{Positive F1} & \textbf{Strict F1} \\
        \midrule
        \emph{Fine-tuning LMs for Spans}              &                      &                    \\
        \quad \emph{Basic Postprocessing}             &                      &                    \\
        \qquad XLM-RoBERTa-Large                      & \textbf{0.891}       & \textbf{0.631}     \\
        \quad \emph{Extended Postprocessing}          &                      &                    \\
        \qquad GBERT-Large                            & \---                 & 0.623              \\
        \midrule
        \emph{Training SVM for Binary Classification} &                      &                    \\
        \quad Qwen3-Embedding-8B                      & 0.875                & \---               \\
        \bottomrule
    \end{tabular}
    \caption{Performance scores on the test set for models submitted to the Shared Task.}
    \label{tab:results_test}
\end{table*}

Among the approaches developed for Subtask 1, the Qwen3~Embedding model performed better than GBERT, with an average positive F1-score of 0.901 compared to 0.887 for GBERT (see lower part of Table \ref{tab:results_val}).
In addition to these approaches, we investigated whether the span detection models trained for Subtask 2 could be leveraged to indirectly identify candy speech comments in Subtask 1.
To this end, we applied the Subtask 2 models to the data, postprocessed the predicted spans, and classified a comment as candy speech if it included at least one detected span of any type.
We observed this approach to slightly outperform the dedicated binary classification models on the validation folds of the cross-validation setup, achieving an average positive F1-score of 0.903 and 0.913 for GBERT and XLM-RoBERTa, respectively (see upper part of Table \ref{tab:results_val}).

Based on the results obtained during model exploration, our Subtask 1 submissions consisted of the Qwen3~Embedding model with the SVM classifier and the predictions derived from the XLM-RoBERTa model fine-tuned for span detection.
Both models were retrained on the entire training dataset and evaluated on the test set by the Shared Task organizers.
Consistent with our validation findings, the XLM-RoBERTa derivate outperformed the Qwen3~Embedding model on the test set, achieving positive F1-scores of 0.891 and 0.875 and earning first and 8th place, respectively (see Table \ref{tab:results_test}).

For Subtask 2, we explored the fine-tuning of GBERT and XLM-RoBERTa model architectures and the set of postprocessing steps used to parse BIO-tokens into spans.
We trained both models on the entire training dataset and evaluated them on the early stopping set, finding that XLM-RoBERTa slightly outperformed GBERT, regardless of the postprocessing steps used.
Interestingly, GBERT seemed to benefit from the more advanced postprocessing steps, slightly increasing the strict F1-score from 0.731 to 0.739, while XLM-RoBERTa's scores declined from 0.747 to 0.742.

For our submissions to Subtask 2, we retrained the GBERT model with the advanced set of postprocessing steps and the XLM-RoBERTa model with the basic set of postprocessing steps.
XLM-RoBERTa achieved a strict F1-score of 0.631 on the test data, earning first place in Subtask 2, while GBERT achieved a strict F1-score of 0.623, earning second place.

\section{Conclusion}
\label{sec:conclusion}

In our study, we demonstrated that even in the era of generalist LLMs, fine-tuning smaller, specialized models remains a competitive strategy for specific tasks, such as the identification of candy speech in German YouTube comments.
Using a single fine-tuned XLM-RoBERTa model, our top submissions to the \emph{GermEval 2025 Shared Task on Candy Speech Detection} secured first place in both Subtasks.
Although the model was trained for span detection in Subtask 2, it was able to identify candy speech comments in Subtask 1 by predicting a comment as containing candy speech whenever a span was detected.
It even outperformed models trained specifically for binary classification in Subtask 1, which we hypothesize is due to the richer training signal provided by token-level (span) annotations, as opposed to the comment-level binary labels of Subtask 1.

We consider several directions as promising to further improve our model's performance: (i) fine-tuning larger models like the Qwen3 Embedding model, (ii) creating ensembles similar to previous GermEval contributions \cite{Bornheim2021,Blaneck2022}, and (iii) expanding the context of each comment (e.g., by including previous comments or video content).
Future work should address our model's current inability to handle overlapping spans, which was a trade-off we accepted for the sake of simplicity in the GermEval 2025 Shared Task, given the low prevalence of overlapping spans in the training data.
This limitation may be overcome by reframing the problem as a multi-label task, as demonstrated by the winning submission to the GermEval 2023 Shared Task \cite{Ehrmanntraut2023}.
Finally, we recommend evaluating our model on data from different media platforms before deploying it in real-world applications.
We anticipate that candy speech detection models, such as those presented in this paper, will be valuable not only in social media moderation but also as essential components of control mechanisms in generative LLMs, where they could help to monitor sycophancy and other forms of user manipulation.

\section*{Acknowledgements}
We are grateful to M. Reißel and V. Sander for providing us with computing resources.


\begin{thebibliography}{19}
\providecommand{\natexlab}[1]{#1}

\bibitem[{Albladi et~al.(2025)Albladi, Islam, Das, Bigonah, Zhang, Jamshidi, Rahgouy, Raychawdhary, Marghitu, and Seals}]{Albladi2025}
Aish Albladi, Minarul Islam, Amit Das, Maryam Bigonah, Zheng Zhang, Fatemeh Jamshidi, Mostafa Rahgouy, Nilanjana Raychawdhary, Daniela Marghitu, and Cheryl~D. Seals. 2025.
\newblock \href {https://doi.org/10.1109/ACCESS.2025.3532397} {Hate Speech Detection Using Large Language Models: {A} Comprehensive Review}.
\newblock \emph{{IEEE} Access}, 13:20871--20892.

\bibitem[{Blaneck et~al.(2022)Blaneck, Bornheim, Grieger, and Bialonski}]{Blaneck2022}
Patrick~Gustav Blaneck, Tobias Bornheim, Niklas Grieger, and Stephan Bialonski. 2022.
\newblock \href {https://aclanthology.org/2022.germeval-1.10} {Automatic Readability Assessment of {G}erman Sentences with Transformer Ensembles}.
\newblock In \emph{Proceedings of the GermEval 2022 Workshop on Text Complexity Assessment of German Text, GermEval@KONVENS 2022}, pages 57--62, Potsdam, Germany. Association for Computational Linguistics.

\bibitem[{Bonaldi et~al.(2024)Bonaldi, Chung, Abercrombie, and Guerini}]{Bonaldi2024}
Helena Bonaldi, Yi{-}Ling Chung, Gavin Abercrombie, and Marco Guerini. 2024.
\newblock \href {https://doi.org/10.18653/V1/2024.FINDINGS-NAACL.221} {{NLP} for Counterspeech against Hate: {A} Survey and How-To Guide}.
\newblock In \emph{Findings of the Association for Computational Linguistics: {NAACL} 2024, Mexico City, Mexico, June 16-21, 2024}, pages 3480--3499. Association for Computational Linguistics.

\bibitem[{Bornheim et~al.(2021)Bornheim, Grieger, and Bialonski}]{Bornheim2021}
Tobias Bornheim, Niklas Grieger, and Stephan Bialonski. 2021.
\newblock \href {https://aclanthology.org/2021.germeval-1.16} {{FHAC} at GermEval 2021: Identifying {G}erman toxic, engaging, and fact-claiming comments with ensemble learning}.
\newblock In \emph{Proceedings of the GermEval 2021 Shared Task on the Identification of Toxic, Engaging, and Fact-Claiming Comments, GermEval@KONVENS 2021}, pages 105--111, D{\"{u}}sseldorf, Germany. Association for Computational Linguistics.

\bibitem[{Bornheim et~al.(2024)Bornheim, Grieger, Blaneck, and Bialonski}]{Bornheim2024}
Tobias Bornheim, Niklas Grieger, Patrick~Gustav Blaneck, and Stephan Bialonski. 2024.
\newblock \href {https://doi.org/10.21248/jlcl.37.2024.244} {Speaker Attribution in {G}erman Parliamentary Debates with {QL}o{RA}-adapted Large Language Models}.
\newblock \emph{Journal for Language Technology and Computational Linguistics}, 37(1):1--13.

\bibitem[{Chakravarthi et~al.(2022)Chakravarthi, Muralidaran, Priyadharshini, Navaneethakrishnan, McCrae, Garc{\'{\i}}a, Jim{\'{e}}nez{-}Zafra, Valencia{-}Garc{\'{\i}}a, Kumaresan, Ponnusamy, Garc{\'{\i}}a{-}Baena, and Garc{\'{\i}}a{-}D{\'{\i}}az}]{Chakravarthi2022b}
Bharathi~Raja Chakravarthi, Vigneshwaran Muralidaran, Ruba Priyadharshini, Subalalitha~Chinnaudayar Navaneethakrishnan, John~P. McCrae, Miguel~{\'{A}}ngel Garc{\'{\i}}a, Salud~Mar{\'{\i}}a Jim{\'{e}}nez{-}Zafra, Rafael Valencia{-}Garc{\'{\i}}a, Prasanna~Kumar Kumaresan, Rahul Ponnusamy, Daniel Garc{\'{\i}}a{-}Baena, and Jos{\'{e}}~Antonio Garc{\'{\i}}a{-}D{\'{\i}}az. 2022.
\newblock \href {https://doi.org/10.18653/V1/2022.LTEDI-1.58} {Overview of the Shared Task on Hope Speech Detection for Equality, Diversity, and Inclusion}.
\newblock In \emph{Proceedings of the Second Workshop on Language Technology for Equality, Diversity and Inclusion, {LT-EDI} 2022, Dublin, Ireland, May 27, 2022}, pages 378--388. Association for Computational Linguistics.

\bibitem[{Chan et~al.(2020)Chan, Schweter, and M{\"{o}}ller}]{Chan2020}
Branden Chan, Stefan Schweter, and Timo M{\"{o}}ller. 2020.
\newblock \href {https://doi.org/10.18653/v1/2020.coling-main.598} {German's Next Language Model}.
\newblock In \emph{Proc. 28th Int. Conf. on Computational Linguistics, {COLING} 2020}, pages 6788--6796, Barcelona, Spain (Online). International Committee on Computational Linguistics.

\bibitem[{Clausen et~al.(2025)Clausen, Scheffler, and Wiegand}]{Clausen2025}
Yulia Clausen, Tatjana Scheffler, and Michael Wiegand. 2025.
\newblock Overview of the {GermEval} 2025 {S}hared {T}ask on {C}andy {S}peech {D}etection.
\newblock In \emph{Proceedings of the 21st Conference on Natural Language Processing ({KONVENS} 2025): Workshops}, Hildesheim, Germany. ACL.

\bibitem[{Conneau et~al.(2020)Conneau, Khandelwal, Goyal, Chaudhary, Wenzek, Guzm{\'{a}}n, Grave, Ott, Zettlemoyer, and Stoyanov}]{Conneau2020}
Alexis Conneau, Kartikay Khandelwal, Naman Goyal, Vishrav Chaudhary, Guillaume Wenzek, Francisco Guzm{\'{a}}n, Edouard Grave, Myle Ott, Luke Zettlemoyer, and Veselin Stoyanov. 2020.
\newblock \href {https://doi.org/10.18653/V1/2020.ACL-MAIN.747} {Unsupervised Cross-lingual Representation Learning at Scale}.
\newblock In \emph{Proceedings of the 58th Annual Meeting of the Association for Computational Linguistics, {ACL} 2020, Online, July 5-10, 2020}, pages 8440--8451. Association for Computational Linguistics.

\bibitem[{DeepSeek{-}AI et~al.(2024)DeepSeek{-}AI, Liu, Feng, Xue, Wang, Wu, Lu, Zhao, Deng, Zhang, Ruan, Dai, Guo, Yang, Chen, Ji, Li, Lin, Dai, Luo, Hao, Chen, Li, Zhang, Bao, Xu, Wang, Zhang, Ding, Xin, Gao, Li, Qu, Cai, Liang, Guo, Ni, Li, Wang, Chen, Chen, Yuan, Qiu, Li, Song, Dong, Hu, Gao, Guan, Huang, Yu, Wang, Zhang, Xu, Xia, Zhao, Wang, Zhang, Li, Wang, Zhang, Zhang, Tang, Li, Tian, Huang, Wang, Zhang, Wang, Zhu, Chen, Du, Chen, Jin, Ge, Zhang, Pan, Wang, Xu, Zhang, Chen, Li, Lu, Zhou, Chen, Wu, Ye, Ye, Ma, Wang, Zhou, Yu, Zhou, Pan, Wang, Yun, Pei, Sun, Xiao, and Zeng}]{DeepSeek2024}
DeepSeek{-}AI, Aixin Liu, Bei Feng, Bing Xue, Bingxuan Wang, Bochao Wu, Chengda Lu, Chenggang Zhao, Chengqi Deng, Chenyu Zhang, Chong Ruan, Damai Dai, Daya Guo, Dejian Yang, Deli Chen, Dongjie Ji, Erhang Li, Fangyun Lin, Fucong Dai, and 81 others. 2024.
\newblock \href {https://doi.org/10.48550/ARXIV.2412.19437} {Deep{S}eek-{V}3 Technical Report}.
\newblock \emph{CoRR}, abs/2412.19437.

\bibitem[{Devlin et~al.(2019)Devlin, Chang, Lee, and Toutanova}]{Devlin2018}
Jacob Devlin, Ming{-}Wei Chang, Kenton Lee, and Kristina Toutanova. 2019.
\newblock \href {https://doi.org/10.18653/v1/n19-1423} {{BERT:} {P}re-training of Deep Bidirectional Transformers for Language Understanding}.
\newblock In \emph{Proc. 2019 Conf. North American Chapter of the Association for Computational Linguistics: Human Language Technologies, {NAACL-HLT} 2019}, volume~1, pages 4171--4186, Minneapolis, MN, USA. Association for Computational Linguistics.

\bibitem[{Ehrmanntraut(2023)}]{Ehrmanntraut2023}
Anton Ehrmanntraut. 2023.
\newblock Politics, {BERTed}: Automatic Attribution of Speech Events in German Parliamentary Debates.
\newblock In \emph{Proceedings of the {GermEval} 2023 Shared Task on Speaker Attribution in Newswire and Parliamentary Debates ({SpkAtt}-2023)}, pages 22--30.

\bibitem[{Geiss and Zarcone(2024)}]{Geiss2024}
Corsin Geiss and Alessandra Zarcone. 2024.
\newblock \href {https://aclanthology.org/2024.germeval-2.2/} {{THA}ugs at {G}erm{E}val 2024 (Shared Task 1: {G}er{MS}-Detect): Predicting the Severity of Misogyny/Sexism in Forum Comments with {BERT} Models (Subtask 1, Closed Track and Additional Experiments)}.
\newblock In \emph{Proceedings of GermEval 2024 Task 1 GerMS-Detect Workshop on Sexism Detection in German Online News Fora (GerMS-Detect 2024)}, pages 10--20, Vienna, Austria. Association for Computational Lingustics.

\bibitem[{Kubin et~al.(2024)Kubin, Merz, Wahba, Davis, Gray, and von Sikorski}]{Kubin2024}
Emily Kubin, Pascal Merz, Mariam Wahba, Cate Davis, Kurt Gray, and Christian von Sikorski. 2024.
\newblock \href {https://doi.org/10.3389/fcomm.2024.1447457} {Understanding news-related user comments and their effects: {A} systematic review}.
\newblock \emph{Frontiers in Communication}, 9:1447457.

\bibitem[{OpenAI(2023)}]{GPT4-2023}
OpenAI. 2023.
\newblock \href {https://doi.org/10.48550/arXiv.2303.08774} {{GPT-4} Technical Report}.
\newblock \emph{CoRR}, abs/2303.08774.

\bibitem[{Ramshaw and Marcus(1995)}]{Ramshaw1995}
Lance~A. Ramshaw and Mitch Marcus. 1995.
\newblock \href {https://aclanthology.org/W95-0107/} {Text Chunking using Transformation-Based Learning}.
\newblock In \emph{Third Workshop on Very Large Corpora, VLC@ACL 1995, Cambridge, Massachusetts, USA, June 30, 1995}.

\bibitem[{Tucudean et~al.(2024)Tucudean, Bucos, Dragulescu, and Caleanu}]{Tucudean2024}
Georgiana Tucudean, Marian Bucos, Bogdan Dragulescu, and Catalin{-}Daniel Caleanu. 2024.
\newblock \href {https://doi.org/10.7717/PEERJ-CS.2222} {Natural language processing with transformers: {A} review}.
\newblock \emph{PeerJ Comput. Sci.}, 10:e2222.

\bibitem[{Yang et~al.(2025)Yang, Li, Yang, Zhang, Hui, Zheng, Yu, Gao, Huang, Lv, Zheng, Liu, Zhou, Huang, Hu, Ge, Wei, Lin, Tang, Yang, Tu, Zhang, Yang, Yang, Zhou, Zhou, Lin, Dang, Bao, Yang, Yu, Deng, Li, Xue, Li, Zhang, Wang, Zhu, Men, Gao, Liu, Luo, Li, Tang, Yin, Ren, Wang, Zhang, Ren, Fan, Su, Zhang, Zhang, Wan, Liu, Wang, Cui, Zhang, Zhou, and Qiu}]{Yang2025}
An~Yang, Anfeng Li, Baosong Yang, Beichen Zhang, Binyuan Hui, Bo~Zheng, Bowen Yu, Chang Gao, Chengen Huang, Chenxu Lv, Chujie Zheng, Dayiheng Liu, Fan Zhou, Fei Huang, Feng Hu, Hao Ge, Haoran Wei, Huan Lin, Jialong Tang, and 41 others. 2025.
\newblock \href {https://doi.org/10.48550/ARXIV.2505.09388} {Qwen3 Technical Report}.
\newblock \emph{CoRR}, abs/2505.09388.

\bibitem[{Zhang et~al.(2025)Zhang, Li, Long, Zhang, Lin, Yang, Xie, Yang, Liu, Lin, Huang, and Zhou}]{Zhang2025}
Yanzhao Zhang, Mingxin Li, Dingkun Long, Xin Zhang, Huan Lin, Baosong Yang, Pengjun Xie, An~Yang, Dayiheng Liu, Junyang Lin, Fei Huang, and Jingren Zhou. 2025.
\newblock \href {https://doi.org/10.48550/ARXIV.2506.05176} {Qwen3 {E}mbedding: {A}dvancing Text Embedding and Reranking Through Foundation Models}.
\newblock \emph{CoRR}, abs/2506.05176.

\end{thebibliography}
\end{document}